\documentclass{article}

\usepackage{arxiv}

\usepackage[utf8]{inputenc} 
\usepackage[T1]{fontenc}    
\usepackage{hyperref}       
\usepackage{url}            
\usepackage{booktabs}       
\usepackage{amsfonts}       
\usepackage{nicefrac}       
\usepackage{microtype}      
\usepackage{lipsum}		
\usepackage{graphicx}
\usepackage{natbib}
\usepackage{doi}
\usepackage{amsmath,amssymb,amsfonts}
\usepackage{algorithmic}
\usepackage{graphicx}
\usepackage{textcomp}
\usepackage{xcolor}

\title{Compact Binary Fingerprint for Image Copy Re-Ranking }

\author{Nazar Mohammad \and Junaid Baber \and Maheen Bakhtyar \and Bilal Ahmed Chandio \and Anwar Ali Sanjrani \\
Department of CS and IT, University of Balochistan.
}

\renewcommand{\shorttitle}{\textit{arXiv} Template}

\hypersetup{
pdftitle={BinarySIFT},
pdfsubject={Image Retrieval},
pdfauthor={Junaid Baber},
pdfkeywords={SIFT Ranking · Binary Feature Space · Image Copy Retrieval},
}

\begin{document}
\maketitle

\begin{abstract}
	Image copy detection is challenging and appealing topic in computer vision and signal processing.  Recent advancements in multimedia have made distribution of image across the global easy and fast: that leads to many other issues such as forgery and image copy retrieval. 

Local keypoint descriptors such as SIFT are used to represent the images, and based on those descriptors matching, images are matched and retrieved. Features are quantized so that searching/matching may be made feasible for large databases at the cost of accuracy loss. In this paper, we propose binary feature that is obtained by quantizing the SIFT into binary, and rank list is re-examined to remove the false positives. Experiments on challenging dataset shows the gain in accuracy and time.
\end{abstract}


\section{Introduction}
Internet is the most powerful tool used in these days. Visual graphics like videos and images are the most attention seekers rather than text paragraphs, so that image manipulation and editing softwares are easily available online or offline. That's why thousands of images are being edited and shared among different servers of social media and other sites such as Facebook\footnote{https://www.facebook.com/}, Flicker\footnote{http://www.flickr.com/}, and ImageShake\footnote{http://imageshack.us/}. Image databases are expanding rapidly in scope and immensity  which is being a major cause of in-efficient image recuperation such as image infringement and extravagant. Many researchers proclaimed different elucidations to fix this problem; image copy retrieval and partial-duplicate image retrieval overcome the problem in some such way. Image copy retrieval and partial-duplicate image retrieval somehow solve these problems. 
Image copy is defined as a segment of image derived 
from another image usually by means of
various challenging transformations such as  pattern addition, content deletion, 
modification of image contents (such as aspect ratio, color, contrast, encoding), cam-cording,
and morphing.  

Local attributes are extensively used for image and video applications. 
Initially, key-points are detected from the images and 
represented by some robust descriptors 
such as SIFT~\cite{David2004}, GLOH~\cite{Mikolajczyk2005}, and CSLBP~\cite{CSLBP}.

The affine regions are decisive in the calculation of the descriptors over the key-points; 
the descriptors performance around the key-points is very delicate while assessment of the affine regions~\cite{Mikolajczyk2005}. 
Therefore, the descriptors computation only prosperous and immense recapitulation affine regions are preferred~\cite{Krystian2004,Mikolajczyk2005}.
The descriptors centered from the affine regions computation to key-points are recapitulation under image transformation, 
but that is not an effective solution because two totally different patches can have same consistency.
The previous research presented about the descriptors have lack in \textit{robustness} and \textit{distinctiveness}~\cite{JICTAbaber}.
The robustness of the descriptors is generally perceived as the limit of descriptors to make due in picture clamor, and unmistakable characterizes the ability of the descriptors to see the difference between two picture fixes or surface. The descriptors processed from various scenes or settings might be viewed as comparative (absence of uniqueness), or comparative descriptors might be considered as different because of image clamor (absence of robustness); the comparability between descriptors is generally registered by the separations between them in include space.

Similarity between pair of images is computed by the distance between the local keypoint descriptors. Since, one image may have several thousand keypoints, so finding the nearest neighbor of single keypoint in  other image, and then repeating the steps for all other keypoints, makes it computationally very expensive. To make searching feasible on large databases, features are quantized. The most common feature quantization is bag of visual word model~(BoVW)~\cite{James07,Zhong2009,Zhou2010, Baber2014}. In BoVW, image is represented by histogram of visual words present in that image, visual words are basically the centroids that are obtained by SIFT features clustering. The BoVW model is computationally efficient compared to raw SIFT based retrieval, whereas, SIFT has better accuracy compared to BoVW. Geometrical connection between key-points and their descriptors are used in the field of image copy detection~\cite{Fergus,LiuCVPR,Satoh2,Satoh1}.
It has been argued that the performance of local features is significantly improved from 
\textit{bag of visual words} to 
\textit{bag of pair of visual words}~\cite{LiuCVPR}. 
Geometrical connection between key-points and their descriptors are used in the field of image copy detection ~\cite{James07,Zhong2009,Zhou2010}. 
Many researchers trying to improve their methods or algorithms to enhance the performance
of content based copy detection (CBCD) performance in the manner of robustness and distinctiveness before the quantization of Bag-of-Visual words~\cite{Xu2011,Ke2004511,Ke04},
or sometimes after BoW quantization~\cite{James07,Zhong2009,Zhou2010}.

In this paper, we propose re-ranking of BoVW by compact binary features. Initially, images are presented by SIFT features, and later quantized into BoVW model. Initial ranking is obtained by computing the distances between query image and databases.  

Furthermore, attempt to improve the heartiness and uniqueness of local key-point descriptors for image duplicate identification sorts of uses. The more robust and distinctive feature descriptors are having a spatial data inside the local patches. Expanding or decreasing the size of the key-points can be helpful to accomplish the goal and makes the descriptor vector more efficient. We tentatively described and shows the presentation of improved CBCD, and the descriptors which are impacted from their region are utilized.

\section{Related work}
\label{sec:related}
This area is divided into two sections. In the first section, we described the different portions of image copy detection and its methods, then the Second section will clarify about two well-known feature descriptors which are used in our experiments.

\subsection{Image copy detection}

Watermarking and CBCD techniques are mostly used to prevent copyright issues. Moreover, CBCD is the correspondent approach for watermarking. Initially, global features were used for CBCD which will later switched into local features.

Chang et al.~\cite{Edward1998} propose RIME (Replicated IMage dEtector) technique through which we can identifies the pilfered duplicate images on the world wide web  by utilizing color space and wavelets~\cite{chang1998rime}.. The fundamental sort of changes can be made with a good accuracy by the system.
Kim~\cite{Kim2003} utilizes Discrete Cosine Transform (DCT) for CBCD, as DCT is increasingly vigorous to numerous mutilations and image changes. Images are changed over into YUV format and \emph{Y} segment is utilized, as it is contended that hues don't have essential influence in image recovery. RIME effectively identifies the duplicates of the test image with and without alterations, anyway they neglect to identify the duplicates with $90^{\circ}$ or $270^{\circ}$ turn. 
The worldwide highlights are effective for basic sorts of changes, be that as it may, if there should arise an occurrence of serious changes, the exhibition of worldwide highlights is poor; for instance, if there should arise an occurrence of trimming, impediment, and angle proportion change. 
There are many scheme proposed for the image copy detection and forgery detection. SIFT and SURF are the mostly used methods for the feature extraction. Cong lin et al.~\cite{lin2019copy} proposed a method that combines SIFT and Local Intensity Order Pattern (LIOP) to adopt good feature results. As SIFT is invariant feature extractor with image scale, image rotation and noise removal etc. Meanwhile LIOP descriptor is used for image scale, image rotation, viewpoint change, image blur and JPEG compression.
 Local features have proven to be more resistant and robust 
for severe image transformations as compared to the global features. 
Local features have demonstrated to be progressively safe and powerful for extreme image changes when contrasted with the worldwide highlights. Numerous CBCD and image recovery frameworks have been proposed dependent on SIFT and other local features ~\cite{Chum2011,Nister2006,James07,Philbin2011,Zhong2009,Xu2011,Zhou2011,Zhou2010}.
Xu et al.~\cite{Xu2011} Local features have demonstrated to be progressively safe and powerful for extreme image changes when contrasted with the worldwide highlights. Numerous CBCD and image recovery frameworks have been proposed dependent on SIFT and other local features.
Zhou et al.~\cite{Zhou2010} propose a system for fractional copy image location for huge scope applications by utilizing sack of-visual-words model. They quantized the SIFT in descriptor space and direction space. They encode the spatial design of keypoints by XMAP and YMAP technique, which assists with expelling the exceptions. Be that as it may, their structure is touchy to computerized blunders, for example, floating or moving of keypoints because of change which causes to miss the genuine matches.
There are different methods are being used for the detection of keypoints and descriptors for instance FAST and its variants \cite{rosten2006machine,rosten2008faster} are used to find keypoints in real-time in which system visual features are matched, like parallel tracking and mapping \cite{klein2007parallel}. It is effective and finds sensible corner keypoints, in the spite of the fact that it must be enlarged pyramid business models for scale \cite{klein2008improving}, and in this case, a Harris corner filter \cite{harris1988combined} is used to dismiss edges and give a sensible score. And for the finding of descriptors, BRIEF \cite{calonder2010brief} is a latest feature descriptor that utilized basic binary tests between pixels in a smoothed image patch. Furthermore, BRIEF is similar with SIFT in many aspects like its robustness, blurriness and perspective distortion. Be that as it may, it is delicate to in-plane rotation. BRIEF as a research method uses binary tests that train a set of classification trees \cite{calonder2008keypoint}. When prepared on a lot of 500 or so run of the typical keypoints, the trees can be used to restore a mark for any self-assertive keypoint \cite{calonder2009high}.

Joly et al.~\cite{joly2007content,joly2003robust} proposed a method which represents 20D spatio-temporal descriptors to each keyframe  that computes about Harris interest points although it requires a large amount of data to store per keyframe, possibly hundreds of Harris key points and their descriptors. 

Recently, image retrieval based on hashing technique has attracted a focus of researchers. Torralba et al.~\cite{torralba2008small} represents a method that can learn short descriptors to retrieve same images from a large database. The technique depends on on a dense 128D global image descriptor, which confines the way to deal with no geometric/perspective invariance. Jain et al.~\cite{jain2008fast} presented a strategy for proficient augmentation of Locally Sensitive Hashing scheme \cite{indyk2000stable} for Mahalanobis distance. Both above mentioned techniques uses a bit string as a unique symbol of the image. In such a portrayal, direct impact of comparative images in a solitary container of the hashing table is impossible and a search over different receptacles must be performed. This is achievable (or indeed, even favorable) for rough closest neighbor or range search when the question model is given. Be that as it may, for grouping undertakings, (for example, discovering all gatherings of close copied pictures in the database) the bit string portrayal is less appropriate.

Over the other low dimensional descriptors, SIFT detector and descriptor gives sufficient performance \cite{mikolajczyk2005performance}. Therefore, it has been used worldwide in the field of copy image retrieval \cite{chum2008near,auclair2009hash}, image classification \cite{nister2006scalable} and processing medical images \cite{jiang2014computer} for instance, tracking the growth of cancerous existence. Several methods have been proposed to accelerate the feature indexing process and reduced the Original SIFT detector's length. And this can be accomplished by ignoring some patches of the original descriptor \cite{khan2011sift}  to get 96D, 64D and 32D descriptors or by employing principle component analysis to obtain 64D SIFT descriptors \cite{ke2004pca}.

Another method for managing worldwide feature extraction is identified with the investigation of data encoded in the image texture. Instances of the calculations that follow this methodology are Steerable Pyramid \cite{simoncelli1995steerable}, Gabor Wavelet Transform \cite{he2002object}, Contour-let Transform \cite{do2005contourlet} or Complex Directional Filter Bank \cite{vo2006texture}. Worldwide highlights calculations are commonly viewed as straightforward and quick, which regularly brings about the absence of in-variance to change of point of view or light. To beat these issues local features techniques were presented \cite{gorecki2012ranking}. For example, Schmid~\cite{schmid1997local}  use Harris corner detector to distinguish intrigue focuses which is insensitive toward change of image direction.

Recki et al.~\cite{gorecki2012ranking} proposed a strategy for The primary commitment of this work is the Ranking by K-means Voting algorithm, whose reason for existing is to make a positioning of comparable pictures. The outcomes got in the trial meeting show the benefit of the technique proposed right now the standard similitude measures, for our situation over the Euclidean distance. Afra et al.~\cite{nurnberger2016near} assessed the exhibition of the RC-SIFT 64D descriptor to tackle the near duplicate recovery task in two cases: Firstly, for benchmarks of various size. Besides, utilizing a similar benchmark yet for various quantities of removed highlights.

Zhihua xu et al.~\cite{xu2011novel} presented a powerful content-based image copy detetion conspire. The fundamental commitment of this paper is in three parts. (1) By looking into the distributed writing about the geometric mutilation strong plans, it is obvious that the nearby invariant areas and worldwide descriptor is an ideal technique to adapt to geometric contortion, especially, editing and revolution; (2) Construct a progression of powerful, homogeneous, and bigger size roundabout patches utilizing the SIFT finder; (3) Adopt the MHD to produce include vector for every round fix.

Wu et al.~\cite{Zhong2009} propose a structure by taking multiple gathering of keypoints rather than single keypoint. They utilize a SIFT keypoints alongside maximally stable extreme areas (MSER) keypoints. MSER keypoints are relative covariant key-points with higher repeatability contrasted with the SIFT key-points, these focuses are likewise bigger in scale and generally littler in number in image. On the off chance that SIFT key-points pack with MSER areas, at the point, they have better robustness and performance with more discrimination power of 45\% accurately precision in Bag-of-Visual Word model as compared with the simple SIFT BoW model. Be that as it may, the computational time for search engine is higher when contrasted with cutting edge. 
Wu et al. also presented a way to deal with locate the close duplicate pictures. The SIFT and MSER feature descriptor were used to locate image duplication. MSER is an area based methodology, while SIFT is a point feature location descriptor. Each of them gathering to turns out the feature extraction to be more distinctive than a single element. The MSER feature extractor is also widely used for the image retrieval system just like SIFT invariant feature extractor. Dissimilar to the SIFT feature detector, MSER recognizes relative covariant stable locales. Each recognized curved locale is standardized into a round district from which a SIFT descriptor is figured out. 

Ahmed Alzu'bi et al.~\cite{alzu2016improving} presents a paper about examination towards discovering smaller and discriminate image portrayals utilizing worldwide and nearby multi-highlight plot. The led tests give bits of knowledge into the connection between image features and other recovery factors, including distance measures, quantization and visual code-books, recovery speed, and memory necessities. A bank of image features is removed and afterward defined into minimized image portrayals. The entirety of the extricated highlights are assessed against eight diverse separation measures for similarity measurements. The exploratory outcomes show that distinctive image features and blends give diverse execution. At the last assessment stage, Euclidean, cosine, and relationship estimates show nearly a similar effect on both retrieval accuracy and efficiency. The Spear-man separation measure has indicated the most noteworthy recovery precision for single neighborhood descriptors contrasted with the joined worldwide or nearby ones. Be that as it may, it takes more coordinating time than other distance measures.
\subsection{Patch based descriptors}

There are three famous approaches for keypoint descriptors calculation.
In the first approach raw pixel values based calculations are used, such as CSLBP and LBP~\cite{CSLBP, LBP1}. 
In the second approach, gradients are computed such as SIFT~\cite{David2004}; and in the third approach 
binary features are used. These binary features can be computed by quantizating the gradient histograms such as BIG-OH and CARD~\cite{Baber2014, CARD}, or sometimes comparing the pixel values, such as BRIEF~\cite{Michael2010}. 
The SIFT is widely used descriptor, and to make this section self-contained, SIFT descriptors is explained below. \\
\textbf{Scale-Invariant Feature Transform} (SIFT)
descriptor is the representation of gradient orientation histograms. SIFT was introduced by David Lowe, after that it was detailed analyzed and deeply developed in 2004~\cite{David2004} to accomplish huge enhancements in stability and feature invariance. The standard SIFT descriptor is built by testing the extents and directions of the image slope in the region around the intrigue focuses, and portraying the significant data of the local region by building smoothed direction histograms\cite{liu2019top}.
To compute the SIFT descriptor,the given patch $P$ 
is divided into grid of $G_x \times G_y$. 
In each cell, the gradient magnitude, $g(x,y)$, 
and orientation, $\theta(x,y)$,  are computed  for each pixel. 
The gradient orientations are divided in to 8 different directions and then histogram of each directions are calculated individually. Each gradient sample added to its histogram by their gradient magnitude and Gaussian weight. 
For Gaussian weight, circular window with a $\sigma$ that is 
1.5 times that of the scale of keypoint  is taken~\cite{David2004}.

\section{Proposed Re-Ranking Framework}
\label{sec:explain}
To improve the accuracy efficiently of BoVW, the obtained rank list is re-evaluated by image to image matching based on binary features. The binary features transform the Euclidean space to binary space that makes the distance computation very fast without significant compromise on accuracy. BoVW model and image to image matching are explained in later in this section. 

\subsection{Image Representation}

Mostly, images are in RGB format which are then presented by gray-scale format, the given image $I\in \mathbb{R}^{m\times n \times c}$, where $m$ and $n$ denote the dimensions (width and height), and $c$ denotes the channels, in case of gray-scale format $c=1$ and image then can be defined as $I\in \mathbb{R}^{m\times n}$. 
Color is most important and huge element of an image if it is maintained in most significant way and in a perceptually situated manner. Content Based Image Retrieval (CBIR) system proposed color as a most significant element in image retrieval. The color histogram was acquainted to distinguish the relationship between color circulation of pixels and spatial relationship of colors \cite{galshetwar2019local}.
Image matching is the method to find out geometric correspondence between the at least two pictures of a similar scene. These images could be of different frames, different time taken and different viewpoints. There are many detectors used for the feature extraction and image matching. We have used Harris-Affine detector for the feature key-point detection. Harris-Affine detector is based on the Harris-Laplace detector, which is a corner detector mostly used in image matching. Phase correlation method is applied to overcome the mismatching issues by estimating the translation parameters of the image and the image which is used to be matched for the feature detection\cite{zheng2019image}. Despite the fact that there is some unobtrusive difference between the corner detection and feature extraction but both are mean to be same in this paper.
We use Harris Affine (HA) keypoint detector~\cite{Krystian2004}.
The HA keypoints have corner like structure with less localization error 
and higher repeatability as compared to Difference of Gaussian (DOG) keypoints 
which is used by SIFT algorithm~\cite{Krystian2004}. 
Each key-point is spoken to by a circular area, which is controlled by scale, slope edge and second grid~\cite{Krystian2004}.
The detected elliptical region is mapped 
to circular region and then normalized into $41 \times 41$ 
pixels in Cartesian grid~\cite{CSLBP,Mikolajczyk2005}.

\subsection{Image-to-Image Matching}
Two images can be matched by computing the distance between their feature vectors, and if the distance is closed enough, then those two images are said to be similar. Rank list is obtained by the distance score received in ascending order. However, image to image matching can be easily modeled if there exists only one feature vector for given image such as gray scale histogram or RGB histogram. 
Whereas, this can not be easily modeled for local keypoints descriptors where one image is represented by set of features. For example, in case of SIFT, image is represented by feature matrix $M \in \mathbb{R}^{a\times b}$, where $a$ indicates the dimensions of one descriptor and $b$ indicates the number of key point. The $b$ varies from image to image, it depended on the contents of the image. On average, there are three thousand to four thousand keypoints per image if difference of Gaussian keypoint detector is used. 
Two images are comparable on the off chance that they have numerous basic coordinating points, or the positioned list is gotten dependent on image-to-image keypoint coordinating. 
Let $Q$ and $R$ be the two images with their local keypoints descriptors $Q_p$ and $R_p$, respectively.
The point pair, $d_q \in Q_p$ and $d_r \in R_p$, , is considered a match if following two conditions meet.

\begin{itemize}
	\item Nearest neighbor condition
		\begin{equation}
			\mathcal{D}(d_q, d_r) = \min_{d_k \in R_p} \mathcal{D}(d_q,d_k)
		\end{equation}
	\item Reliable match
			\begin{equation}
				\mathcal{D}(d_q, d_r) < \min_{d_s \in R_p, s\neq k} \mathcal{D}(d_q,d_s) \times \mathcal{S}
			\end{equation}
\end{itemize}

Where $\mathcal{D}$ is some distance estimated and $\mathcal{S}$ is the threshold, which is being utilized for the stable coordinating under commotion conditions, as recommended by David~\cite{David2004}.

The feature distance used for the matching algorithms is an arbitrary, which can be measured by any distance matching models. There is one distance calculating model used widely to compute the distance matching in the field of computer vision is Euclidean distance, which is presented mathematically as: 
  \begin{equation} \label{eq:Euclidean}
\mathcal{D}(d_q,d_r) = \sqrt{\sum_{i=1}^{a}(d_q(i) - d_r(i))^2}
\end{equation}
Where $a$ represents the dimensions of the vector, i.e., 128 in case of SIFT.

\subsection{Bag of Visual Word Model}

The computational cost of image-to-image matching is very high as SIFT is high dimensional vector. 
For image representation Bag-of-Visual-Words (BoVW) model gave an extensive rise to capturing the local pattern variations. Basically the BoVW model is famous in text retrieval from different domains and have great efficiency of providing accurate results. It is generally motivated by Bag-of-Words (BoW) model. Typically a content documents, reports and so on have part of significant words a that can be spoken to by an feature vector for different word checks, Similarly, a BoVW model speak to the image retrieval approach is utilized to portrayed an image and its number of occurrences in different visual words\cite{kulkarni2019comparing}. The computational cost for each key-point from the first image to its nearest neighbor in the other image is very expensive. Moderating the databases can be used by image-to-image matching models. Bag of Visual Word (BoVW) model is used widely for the accurate and feasible searches results in the large databases. 

The BoVW model, $\bold{B}$, is basically the quantizer which quantizes the descriptor $d \in \mathbb{R}^b$

\begin{equation} \label{eq4}
\begin{split}
\bold{B} & : \mathbb{R}^{b} \rightarrow [1,K] \\ 
  & d \rightarrow \bold{B}(d) 
\end{split}
\end{equation}
The $\bold{B}$ quantizes descriptor $d \in \mathbb{R}^{b}$ to any of the $K$ integer cluster index, known as visual word. Visual words are quantized of different local key-points where the histogram of different visual words are calculated from the large image datasets. In this situation, they suggested to keep the value of $K$ greater. In case its keeps the smaller value it will give robust result but not in particular way and in the opposite case where $K$ is even larger than it makes the model more particular but less robust. In the current days, researchers have used number of values lies between 1 million to 1.5 million. \cite{amato2016reducing, philbin2007object}. In the BoVW model the dataset of 2 to 3 thousand images contains approximately 1.0 to 1.5 million vocabulary keypoints makes the model very inadequate \cite{amato2016reducing,philbin2007object}.  

Flat K-mean~\cite{sivic2003video} or hierarchical K-mean~\cite{nister2006scalable} clustering are commonly used to learn the vocabulary of visual words.

  \begin{figure}
    \centering
    \begin{tabular}{c}
      \includegraphics[width=0.6\textwidth]{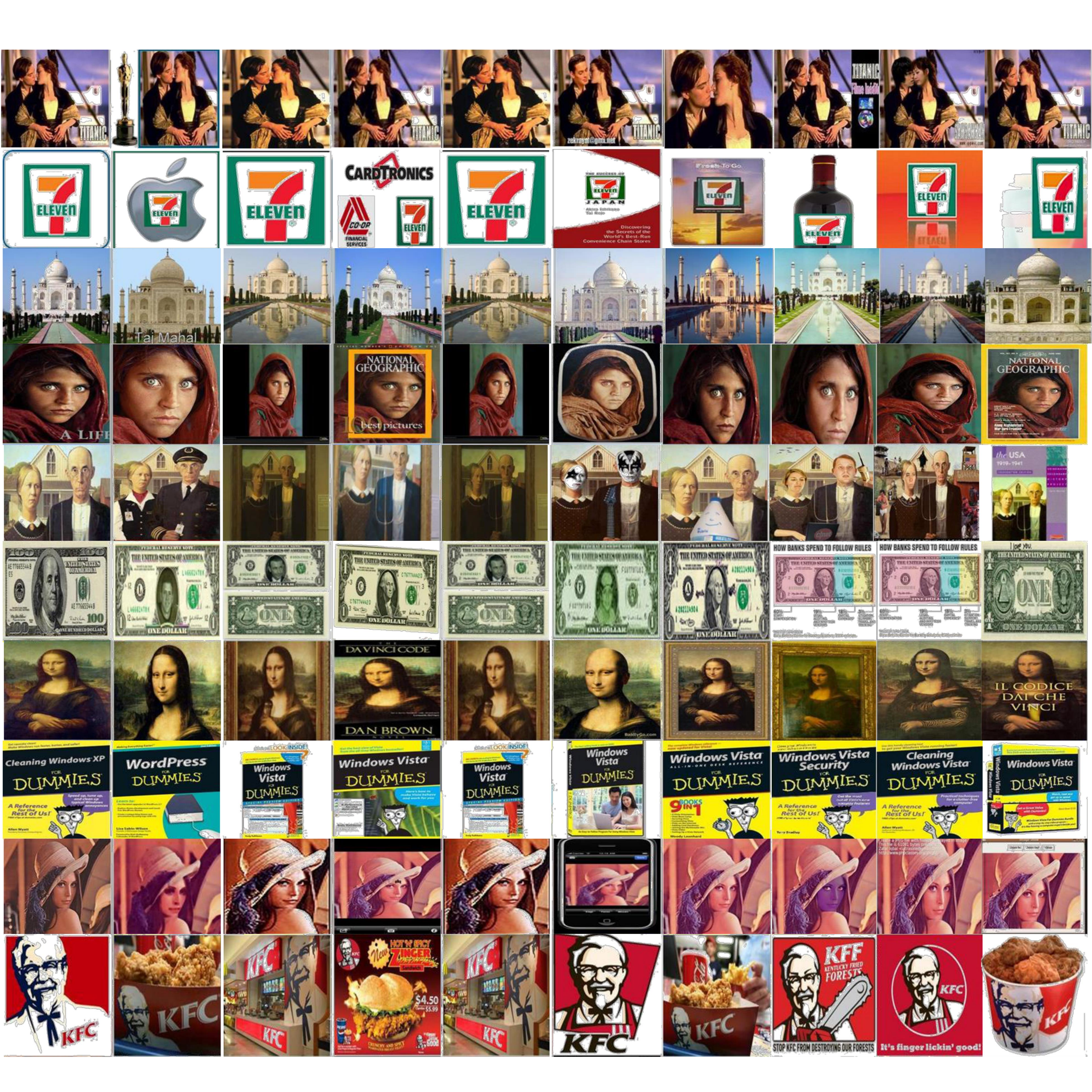}\\      (a) \\
      \includegraphics[width=0.6\textwidth]{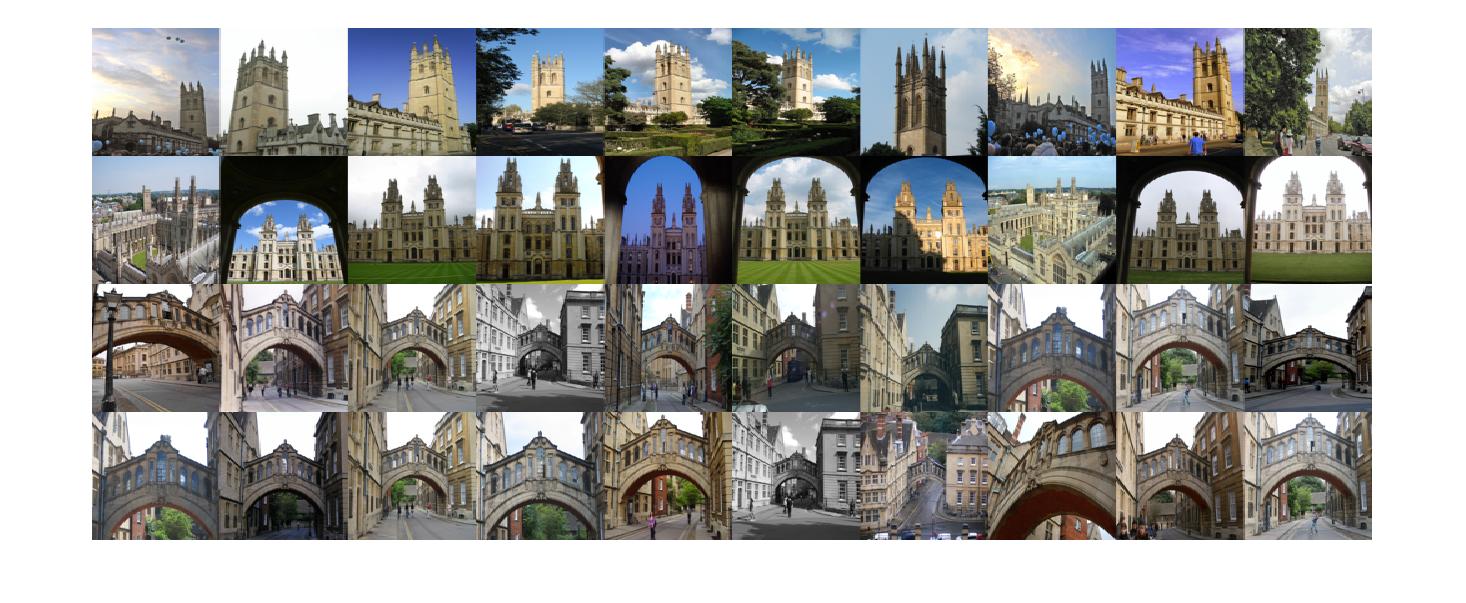} \\         (b) \\
    \end{tabular}
    \caption{CBCD and Oxford-5K dataset samples. (a) shows the CBCD dataset sample of 10 different scenes with their 10 copies, and (b) shows the Oxford-5K dataset sample of 4 landmark with their copies. 
   }
  \label{fig:dataset3}
  \end{figure}

\subsection{Compact Binary Fingerprint: BiSIFT}
Finding the similar images using SIFT is very expensive task, as the the descriptors contains floating values. SIFT is not only computationally expensive for finding the similar images but also storage intensive. We quantize the SIFT into binary vector, Binary SIFT (BiSIFT), similar to BIG-OH, and then compute the image similarity.
The BIG-OH binarizes each gradient histogram on the spatial grid of $4 \times 4$ and then concatenated at the end, whereas, BiSIFT binarizes the whole vector at once. 
Experiments show that there is negligible lose in accuracy but significant gain in computation. For given SIFT descriptor $d\in \mathbb{R}^b$, that can be described as $d = \{ f_1, f_2, \ldots, f_b \}$, where $b=128$, the $d_B\in \mathbb{B}^{b-1}$ is obtained which is $d_B = \{ f^B_1, f^B_2, \ldots, f^B_{b-1} \}$ by 
\begin{equation}
\begin{array}{l}
    f_{i}^B = \left\{
      \begin{array}{l l}
        1 & \quad \text{if} \;\; f_{i}^B \ge f_{i+1}^B\\
        0 & \quad \text{otherwise}.\\
      \end{array} \right.
  \end{array}
\end{equation}
where $i \in \{1,2,\ldots,b-1 \}$. The $d_B$ is 128-Bits binary vector, and it is equivalently efficient as BIG-OH but comparatively simple. 

\section{Experimental Analysis}
This section presents the dataset used for the experiments and evaluation metrics followed by the experimental results. 
\subsection{Datasets}

Three different datasets are used for the evaluation. The first dataset is Synthetic dataset which is used to show the efficiency of the descriptors for similarity computation and storage. The second and third dataset are used to the show effectiveness of proposed binary descriptor.

\subsubsection{Synthetic Dataset}

For the computational speedup tests, we utilize engineered information comprising of uniform irregular bytes for all descriptors, since genuine information isn't expected to assess crude match score figuring execution. This dataset is used only to assure the distance calculation of the raw descriptor about the total required time for the execution. Two different Synthetic descriptors are generated, the first Synthetic descriptors are in floating points, and second Synthetic descriptors are binary descriptors.

\begin{figure*}
  \centering
  \begin{tabular}{cc}
    \includegraphics[width=0.43\textwidth]{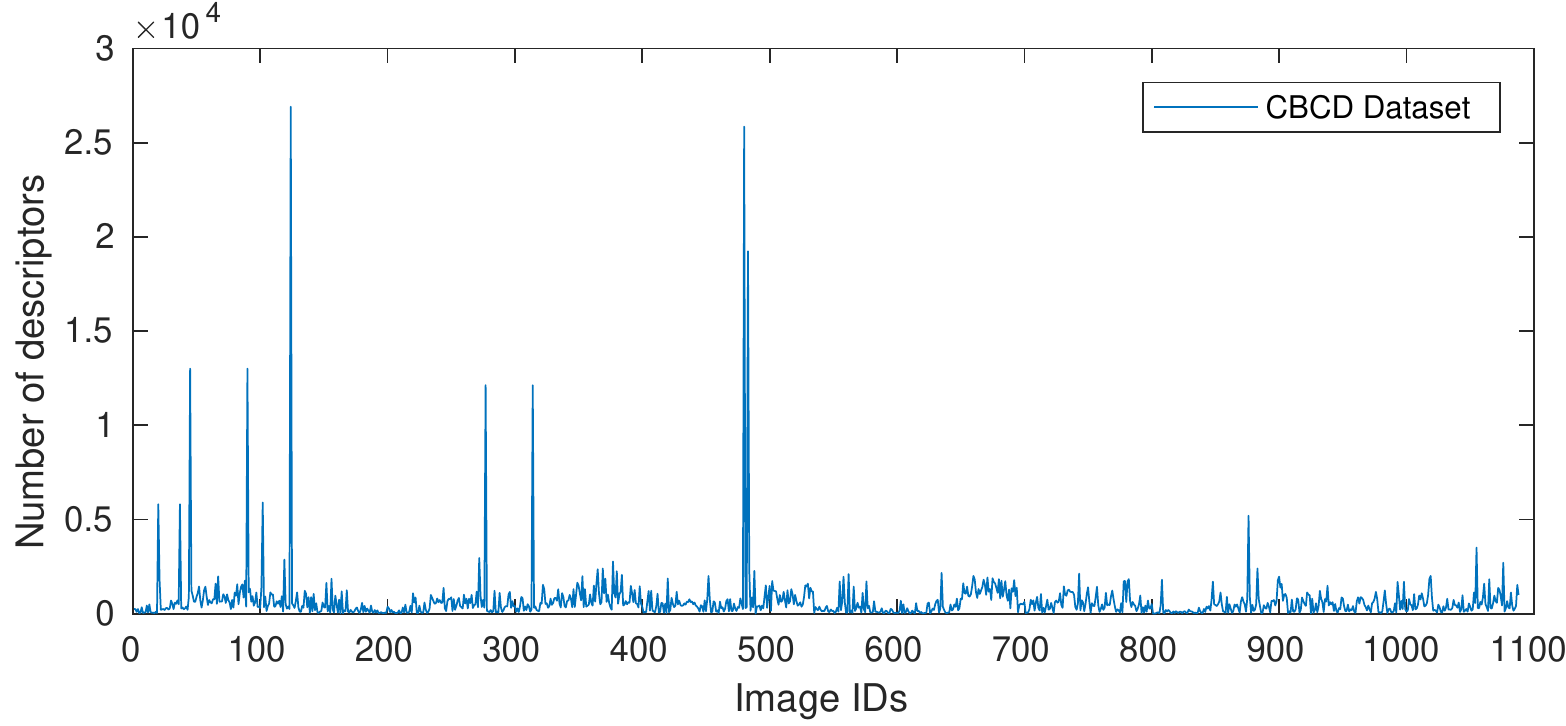}&
    \includegraphics[width=0.45\textwidth]{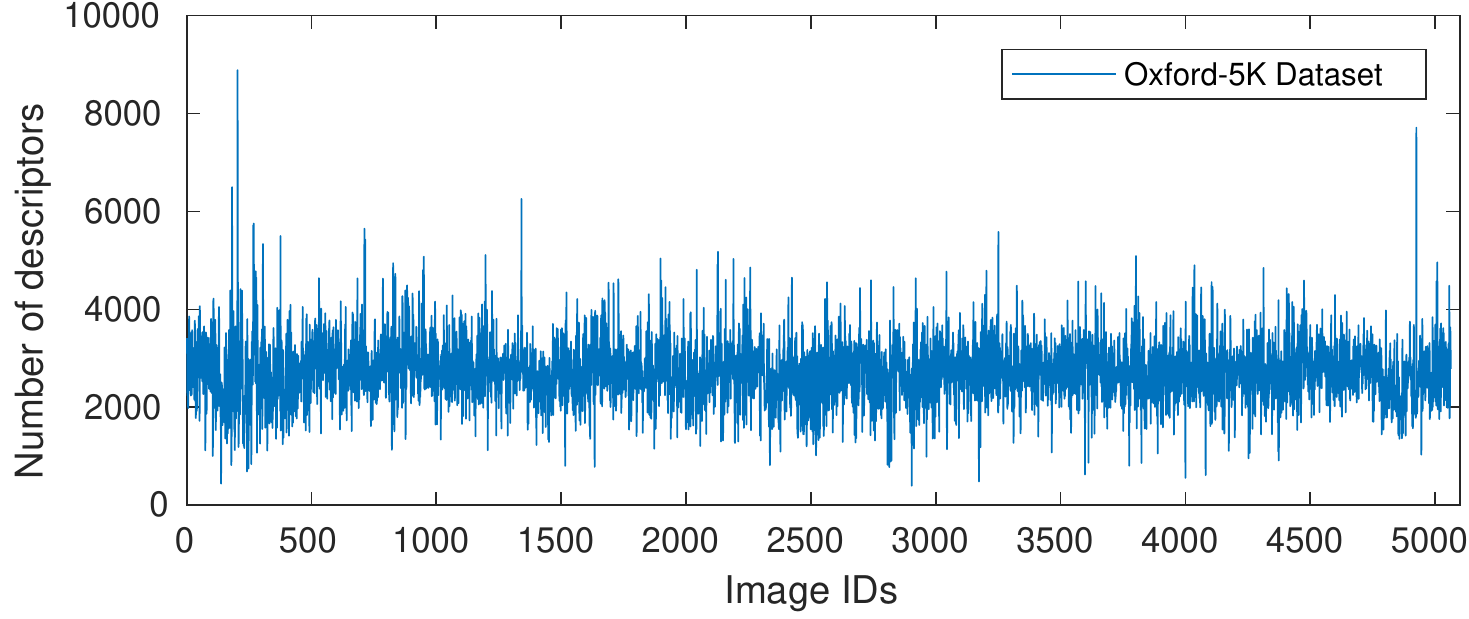} \\(a) & (b)
  \end{tabular}
  \caption{Number of descriptors per image on each dataset. There are total 745563 number key points together and 682.26 on average on CBCD dataset and 13800780 key points together and 2726 on average on the Oxford-5K dataset.
 }
\label{fig:datasets}
\end{figure*}

\subsubsection{Copy Detection Dataset}
The second dataset, CBCD, is provided by Zhou et
al.~\cite{Zhou2010}.  This dataset is taken about 36 different camera scenes with total 1088 number of images. The pictures are seriously misshaped by testing changes normally experienced in duplicate location situation. The changes were made using the camcorder by playing a video of one pictures on the screen but it which was an image inside another image.
The more transformation of an image like adding different designs, JPEG compressions, change of brightness, editing, obscuring, picture flipping, content inclusion, zoom in or out, change of viewpoint and decreasing of an image quality. All of these changes occurring alongside of the moving picture, differentiate changing and picture twisting. Figure~\ref{fig:dataset3}(a) gives one picture haphazardly chose from every scene with their duplicates. 

The third dataset is the Oxford-5K dataset~\cite{James07}. This is the mostly used dataset for the image recovery assessment test. It includes 55 questions pictures of 11 tourists spots including an aggregate of 5063 images gathered from Flickr\footnote{http://www.flickr.com/}, Figure~\ref{fig:dataset3}(b) shows the 4 landmark with their copies.

We have also used an unlabeled dataset of the Paris 100k dataset. This dataset is used for BoVW model learning. Philbin et at.~\cite{James07} explained in detail about these datasets and where it can be found with the original work details of Oxford 5K and Paris 100K.

The visualization of CBCD and Oxford-5K along with description of number of images can be found in  Figure~\ref{fig:datasets}.

\begin{figure}
  \centering
  \begin{tabular}{cc}
    \includegraphics[width=0.43\textwidth]{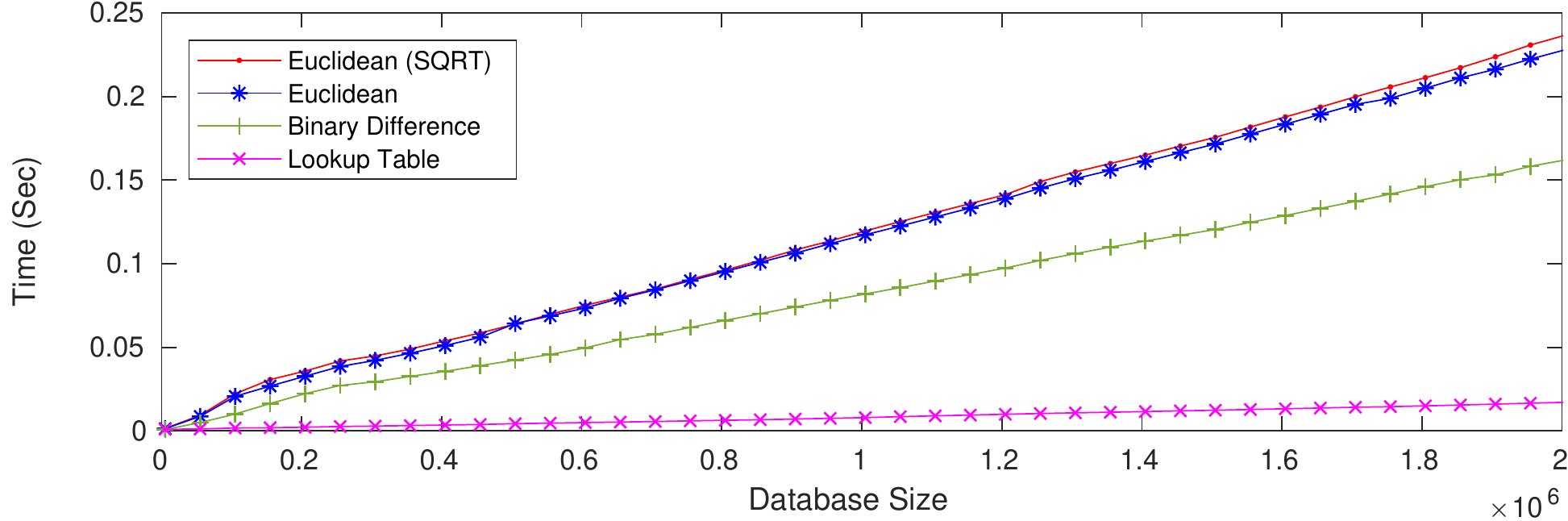}&
    \includegraphics[width=0.45\textwidth]{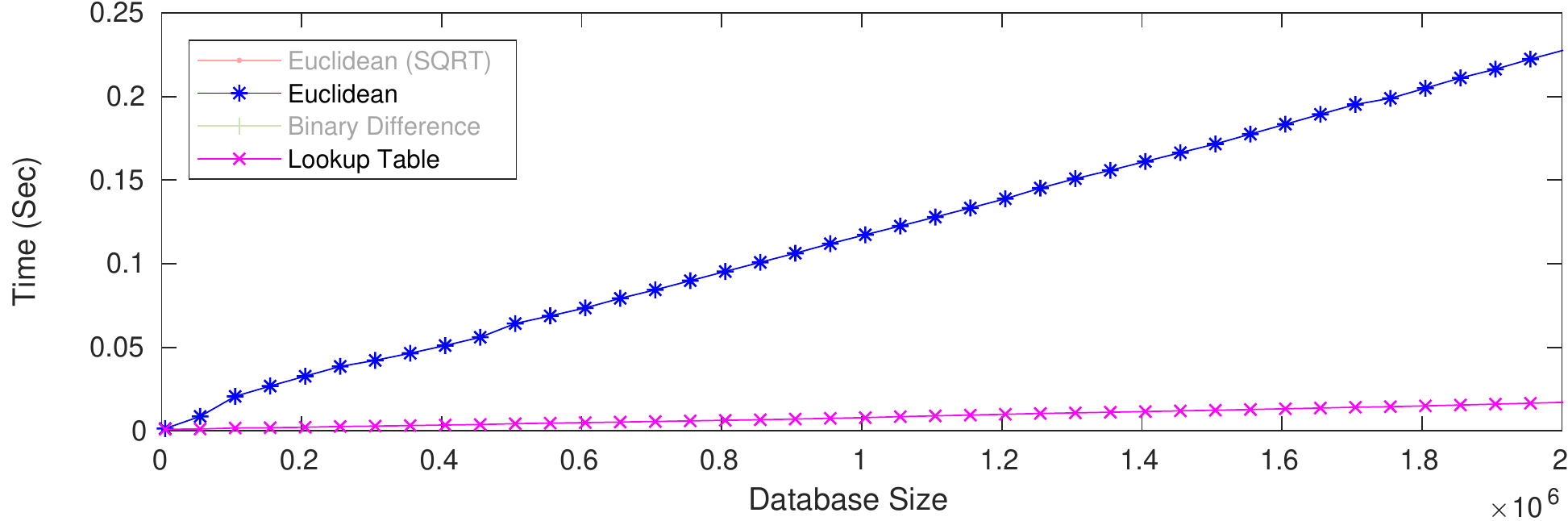} \\(a) & (b)
  \end{tabular}
  \caption{Computational gain to compute the nearest neighbor for one descriptor in varying database size. (a) shows the different possibilities to compute the distance of integer SIFT and BiSIFT, and (b) shows the actual gain of computation between integer SIFT and BiSIFT.
 }
\label{fig:time}
\end{figure}
\subsection{Experiment on Synthetic Dataset}
Synthetic dataset is used to illustrate the computational gain for finding the nearest neighbor (NN) of one keypoint descriptor in large dataset. 
The original SIFT is floating point value that requires 512 bytes/ descriptor~\cite{David2004}. There are number of different implementation of SIFT, VLFEAT~\footnote{https://www.vlfeat.org/} is one of them. Using VLFEAT API, SIFT requires 128~Bytes/ descriptor as each value is 8-Bit unsigned integer. The 128-Bytes implementation of SIFT is very fast compared to original SIFT implementation at the cost of little accuracy loss. The original SIFT takes 252 Seconds to find the NN of given keypoints from the descriptor size of 500K points, whereas, integer implementation takes only 0.04 seconds. The hamming space is further far faster than the integer Euclidean space. That is why original SIFT is quantized into binary, known as BiSIFT, to make NN searching faster. Figure~\ref{fig:time} shows the computational gain of BiSIFT.  In Figure~\ref{fig:time}, the first two curves are the variants of equation~\ref{eq:Euclidean}, the third curve is the sum of the differences of binary vectors, and the last curve shows the lookup based implementation of finding the distances, the idea of lookup based search is inspired by BIG-OH~\cite{Baber2014}.

\subsection{Experiment on Information Retrieval}

This Experiment evaluate the effectiveness of the BiSIFT. As stated earlier, the BiSIFT achieved competitive performance with significant computational gain, the computational gain is shown in the Figure~\ref{fig:time}. Two datasets, CBCD and Oxford-5K, are taken for the experiments. Precision, recall, and mean average Precision (mAP) are used as evaluation matrics for this experiment.

 \textbf{Precision ($P$)}: Precision computes the ratio between correctly retrieved and total retrieved images. The correctly retrieved are basically the true positives (TP), and total retrieved is the sum of true positives and incorrectly retrieved. The incorrectly retrieved are basically the false positives (FP).  It can be computed as follow:
	
	\begin{equation}
	P
	= \dfrac{\text{TP}}{\text{TP}+\text{FP}}
	\label{eq:pr}
	\end{equation}
	
	 \textbf{Recall ($R$)}: Recall computes the ratio between correctly retrieved and total correct. The total correct refers to the total possible true positives of given query image should have retrieved, that can be model as TP + FN, where FN indicates the false negative (images could not be retrieved). It can be computed as follow:
	\begin{equation}
	R
	= \dfrac{\text{TP}}{\text{TP}+\text{FN}} 
	\label{eq:rc}
	\end{equation}
	
	 \textbf{mean Average Precision (mAP)}: An average precision is computed for each theme, each theme has more than one queries, and finally average of the average is calculated, denoted as mAP~\cite{James07}.

\begin{figure}
     \centering
     \includegraphics[width=0.8\textwidth]{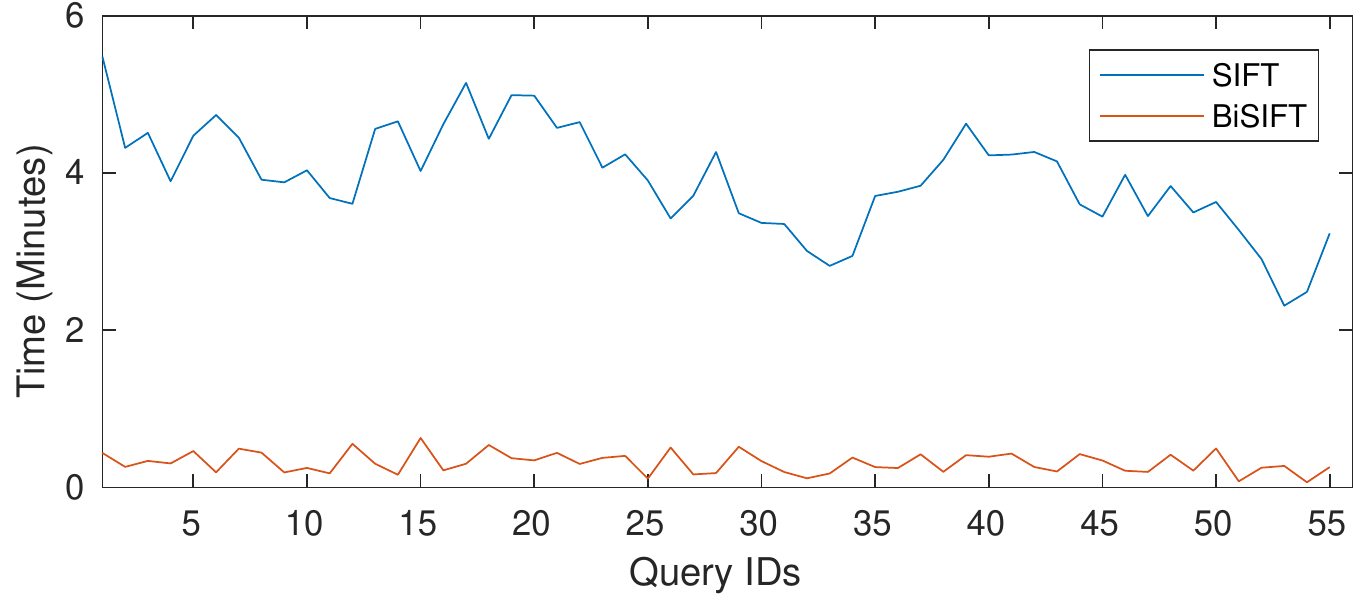}
         \caption{Time to search each query in Oxford-5K dataset. There are 55 queries in Oxford-5K dataset.}
         \label{fig:queryTime}
    \end{figure}

	 Table~\ref{tab:results} show the mAP of BiSIFT along with SIFT, BIG-OH, and BoVW models. In BoVW mode, vocabulary of size 100K is learned on Paris~100K dataset dataset using k-mean clustering. It can be seen that original SIFT is better than BoVW, but BoVW is far efficient compared to the SIFT. On Oxford-5K dataset, average time to search one query image takes 4 minutes on Dell server with 32 GB of RAM.   Figure~\ref{fig:queryTime} shows the time of each query taken on Oxford-5K dataset to be searched in 5K images. Since, every image have different number of keypoints, therefore, the time for each query is different. Whereas, in case of BoVW model, each image is represented by single feature: normalized histogram of visual words present in the image. The BoVW model takes only 0.9 seconds for one query image to search. The feature size of BoVW is $100000 \times 5063$, where rows indicates the vocabulary size and column indicates the number of images.
	 
	 It can be seen that in Table~\ref{tab:results} that the accuracy using BoVW is decreased. To improve the accuracy of BoVW, spatial verification is applied on top $X$ images in the rank-list either by using RANSAC or other techniques~\cite{James07,Zhou2010}. Instead of applying geometric or spatial verification, image to image matching is applied on $X$ images. Since, image to image is more effective but computationally expensive, therefore, it should not be applied on whole dataset but only on $X$ top images on the rank-list, where $X=30$ in the experiments. Image to image based varification improves the accuracy of BoVW with little more time of overhead. It takes only 0.1 seconds using BiSIFT and 1.27 seconds using SIFT, it can be seen that the accuracy is improved when BoVW is piped with SIFT or BiSIFT.

\begin{table}
 \caption{mAP of BiSIFT and SIFT}

\begin{tabular}{c|c|c|c|c|c|c|c|}
\cline{2-8}
 & BoVW & \begin{tabular}[c]{@{}c@{}}BoVW +\\ SIFT\end{tabular} & \begin{tabular}[c]{@{}c@{}}BoVW + \\ BiSIFT\end{tabular} & \begin{tabular}[c]{@{}c@{}}BoVW +\\ BIG-OH\end{tabular} & SIFT & BiSIFT & BIG-OH \\ \hline
\multicolumn{1}{|c|}{CBCD} & 0.47 & 0.57 & 0.58 & 0.60 & 0.61 & 0.59 & 0.60 \\ \hline
\multicolumn{1}{|c|}{Oxford-5K} & 0.51 & 0.53 & 0.52 & 0.53 & 0.55 & 0.52 & 0.53 \\ \hline
\end{tabular}

\label{tab:results}
\end{table}

\section{Conclusion}

Image retrieval is challenging task in large databases as images are easily distorted, modified, and forged. To retrieve the copies or similar images, images are represented by some robust and distinctive local descriptors such as SIFT. However,  local descriptor based image matching and then retrieving is very expensive task. As shown in the experiments, it takes on average 4 minutes to search one query image on the database of size as long as 5K, where feature extraction time is not include: only feature matching and sorting time are included. To make search feasible in real-time, features are quantizes into smaller features space at the cost of accuracy. 
BoVW is widely used in image retrieval for feature quantization. It takes only 0.9 seconds to a query image on the database of size 5K using BoVW model with vocabulary size of 100K. 
In this paper, we quantize the SIFT into binary vector as binary space is faster compared to Euclidean space (Integer values) for searching, as shown in Figure~\ref{fig:time}, without compromising on the accuracy, as shown in Table~\ref{tab:results}.
\section*{Conflict of interest}
The authors declare that they have no conflict of interest.

\bibliographystyle{unsrtnat}
\bibliography{reference2,references_var,references}  

\begin{thebibliography}{63}
\providecommand{\natexlab}[1]{#1}
\providecommand{\url}[1]{\texttt{#1}}
\expandafter\ifx\csname urlstyle\endcsname\relax
  \providecommand{\doi}[1]{doi: #1}\else
  \providecommand{\doi}{doi: \begingroup \urlstyle{rm}\Url}\fi

\bibitem[Lowe(2004)]{David2004}
D.~G. Lowe.
\newblock Distinctive image features from scale-invariant keypoints.
\newblock \emph{IJCV}, 2004.

\bibitem[Mikolajczyk and Schmid(2005{\natexlab{a}})]{Mikolajczyk2005}
K.~Mikolajczyk and C.~Schmid.
\newblock A performance evaluation of local descriptors.
\newblock \emph{IEEE Trans. on PAMI}, 2005{\natexlab{a}}.

\bibitem[Heikkila et~al.(2009)Heikkila, Pietikainen, and Schmid]{CSLBP}
M.~Heikkila, M.~Pietikainen, and C.~Schmid.
\newblock Description of interest regions with local binary patterns.
\newblock \emph{Pattern Recognition}, 2009.

\bibitem[Mikolajczyk and Schmid(2004)]{Krystian2004}
K.~Mikolajczyk and C.~Schmid.
\newblock Scale and affine invariant interest point detectors.
\newblock \emph{IJCV}, 2004.

\bibitem[Baber et~al.(2015)Baber, Fida, Bakhtyar, and Ashraf]{JICTAbaber}
Junaid Baber, Erum Fida, Maheen Bakhtyar, and Humaira Ashraf.
\newblock Making patch based descriptors more distinguishable and robust for
  image copy retrieval.
\newblock In \emph{2015 International Conference on Digital Image Computing:
  Techniques and Applications (DICTA)}, pages 1--8. IEEE, 2015.

\bibitem[Philbin et~al.(2007{\natexlab{a}})Philbin, Chum, Isard, Sivic, and
  Zisserman]{James07}
James Philbin, Ondrej Chum, Michael Isard, Josef Sivic, and Andrew Zisserman.
\newblock {Object retrieval with large vocabularies and fast spatial matching}.
\newblock In \emph{CVPR}, 2007{\natexlab{a}}.

\bibitem[Wu et~al.(2009)Wu, Ke, Isard, and Sun]{Zhong2009}
Zhong Wu, Qifa Ke, Michael Isard, and Jian Sun.
\newblock {Bundling features for large scale partial-duplicate web image
  search}.
\newblock In \emph{CVPR}, 2009.

\bibitem[Zhou et~al.(2010)Zhou, Lu, Li, Song, and Tian]{Zhou2010}
Wengang Zhou, Yijuan Lu, Houqiang Li, Yibing Song, and Qi~Tian.
\newblock Spatial coding for large scale partial-duplicate web image search.
\newblock In \emph{ICMR}, 2010.

\bibitem[Baber. et~al.(2014)Baber., Dailey, Satoh, and Bakhtyar]{Baber2014}
J.~Baber., M.~Dailey, S.~Satoh, and M.~Bakhtyar.
\newblock {BIG-OH:} {BI}narization of {g}radient {o}rientation {h}istograms.
\newblock \emph{Image and Vision Computing}, 2014.

\bibitem[Fergus et~al.(2003)Fergus, Perona, and Zisserman]{Fergus}
R.~Fergus, P.~Perona, and A.~Zisserman.
\newblock Object class recognition by unsupervised scale-invariant learning.
\newblock In \emph{CVPR}, 2003.

\bibitem[Liu et~al.(2008)Liu, Hua, Viola, and Chen]{LiuCVPR}
D.~Liu, Gang Hua, P.~Viola, and Tsuhan Chen.
\newblock Integrated feature selection and higher-order spatial feature
  extraction for object categorization.
\newblock In \emph{CVPR}, 2008.

\bibitem[Morioka and Satoh(2010)]{Satoh2}
N.~Morioka and S.~Satoh.
\newblock Building compact local pairwise codebook with joint feature space
  clustering.
\newblock In \emph{ECCV}, 2010.

\bibitem[Morioka and Satoh(2011)]{Satoh1}
N.~Morioka and S.~Satoh.
\newblock Compact correlation coding for visual object categorization.
\newblock In \emph{ICCV}, 2011.

\bibitem[Xu et~al.(2011{\natexlab{a}})Xu, Ling, Zou, Lu, and Li]{Xu2011}
Zhihua Xu, Hefei Ling, Fuhao Zou, Zhengding Lu, and Ping Li.
\newblock A novel image copy detection scheme based on the local
  multi-resolution histogram descriptor.
\newblock \emph{Multimedia Tools and Applications}, 2011{\natexlab{a}}.

\bibitem[Ke and Sukthankar(2004{\natexlab{a}})]{Ke2004511}
Y.~Ke and R.~Sukthankar.
\newblock Pca-sift: A more distinctive representation for local image
  descriptors.
\newblock In \emph{Proc. CVPR}, pages 511--517, 2004{\natexlab{a}}.

\bibitem[Ke et~al.(2004)Ke, Sukthankar, Huston, Ke, and Sukthankar]{Ke04}
Yan Ke, Rahul Sukthankar, Larry Huston, Yan Ke, and Rahul Sukthankar.
\newblock Efficient near-duplicate detection and sub-image retrieval.
\newblock In \emph{In ACM Multimedia}, 2004.

\bibitem[Chang et~al.(1998{\natexlab{a}})Chang, Wang, Li, and
  Wiederhold]{Edward1998}
Edward~Y. Chang, James~Ze Wang, Chen Li, and Gio Wiederhold.
\newblock {RIME: A Replicated Image Detector for the World-Wide Web}.
\newblock In \emph{Storage and Retrieval for Image and Video Databases},
  1998{\natexlab{a}}.

\bibitem[Chang et~al.(1998{\natexlab{b}})Chang, Wang, Li, and
  Wiederhold]{chang1998rime}
Edward~Y Chang, James~Ze Wang, Chen Li, and Gio Wiederhold.
\newblock Rime: A replicated image detector for the world wide web.
\newblock In \emph{Multimedia Storage and Archiving Systems III}, volume 3527,
  pages 58--67. International Society for Optics and Photonics,
  1998{\natexlab{b}}.

\bibitem[Kim(2003)]{Kim2003}
C.~Kim.
\newblock Content-based image copy detection.
\newblock \emph{Signal Processing: Image Communication}, 2003.

\bibitem[Lin et~al.(2019)Lin, Lu, Huang, Liu, Sun, Lin, and Tan]{lin2019copy}
Cong Lin, Wei Lu, Xinchao Huang, Ke~Liu, Wei Sun, Hanhui Lin, and Zhiyuan Tan.
\newblock Copy-move forgery detection using combined features and transitive
  matching.
\newblock \emph{Multimedia Tools and Applications}, 78\penalty0 (21):\penalty0
  30081--30096, 2019.

\bibitem[Chum et~al.(2011)Chum, Philbin, and Zisserman]{Chum2011}
O.~Chum, J.~Philbin, and A.~Zisserman.
\newblock Near duplicate image detection: Min-hash and tf-idf weighting.
\newblock \emph{Proc. BMVC, 2008}, 2011.

\bibitem[Nist{\'e}r and Stew{\'e}nius(2006)]{Nister2006}
D.~Nist{\'e}r and H.~Stew{\'e}nius.
\newblock Scalable recognition with a vocabulary tree.
\newblock In \emph{CVPR}, 2006.

\bibitem[Philbin et~al.(2008)Philbin, Chum, Isard, Sivic, and
  Zisserman]{Philbin2011}
J.~Philbin, O.~Chum, M.~Isard, J.~Sivic, and A.~Zisserman.
\newblock Lost in quantization: Improving particular object retrieval in large
  scale image databases.
\newblock \emph{CVPR}, 2008.

\bibitem[Zhou et~al.(2011)Zhou, Li, Lu, and Tian]{Zhou2011}
W.~Zhou, H.~Li, Y.~Lu, and Q.~Tian.
\newblock Large scale partial-duplicate image retrieval with bi-space
  quantization and geometric consistency.
\newblock \emph{Proc. ICASSP, 2010}, 2011.

\bibitem[Rosten and Drummond(2006)]{rosten2006machine}
Edward Rosten and Tom Drummond.
\newblock Machine learning for high-speed corner detection.
\newblock In \emph{European conference on computer vision}, pages 430--443.
  Springer, 2006.

\bibitem[Rosten et~al.(2008)Rosten, Porter, and Drummond]{rosten2008faster}
Edward Rosten, Reid Porter, and Tom Drummond.
\newblock Faster and better: A machine learning approach to corner detection.
\newblock \emph{IEEE transactions on pattern analysis and machine
  intelligence}, 32\penalty0 (1):\penalty0 105--119, 2008.

\bibitem[Klein and Murray(2007)]{klein2007parallel}
Georg Klein and David Murray.
\newblock Parallel tracking and mapping for small ar workspaces.
\newblock In \emph{2007 6th IEEE and ACM international symposium on mixed and
  augmented reality}, pages 225--234. IEEE, 2007.

\bibitem[Klein and Murray(2008)]{klein2008improving}
Georg Klein and David Murray.
\newblock Improving the agility of keyframe-based slam.
\newblock In \emph{European conference on computer vision}, pages 802--815.
  Springer, 2008.

\bibitem[Harris et~al.(1988)Harris, Stephens, et~al.]{harris1988combined}
Christopher~G Harris, Mike Stephens, et~al.
\newblock A combined corner and edge detector.
\newblock In \emph{Alvey vision conference}, volume~15, pages 10--5244.
  Citeseer, 1988.

\bibitem[Calonder et~al.(2010)Calonder, Lepetit, Strecha, and
  Fua]{calonder2010brief}
Michael Calonder, Vincent Lepetit, Christoph Strecha, and Pascal Fua.
\newblock Brief: Binary robust independent elementary features.
\newblock In \emph{European conference on computer vision}, pages 778--792.
  Springer, 2010.

\bibitem[Calonder et~al.(2008)Calonder, Lepetit, and Fua]{calonder2008keypoint}
Michael Calonder, Vincent Lepetit, and Pascal Fua.
\newblock Keypoint signatures for fast learning and recognition.
\newblock In \emph{European conference on computer vision}, pages 58--71.
  Springer, 2008.

\bibitem[Calonder et~al.(2009)Calonder, Lepetit, Konolige, Mihelich, and
  Fua]{calonder2009high}
M~Calonder, V~Lepetit, K~Konolige, P~Mihelich, and P~Fua.
\newblock High-speed keypoint description and matching using dense signatures.
\newblock \emph{Under review}, 2, 2009.

\bibitem[Joly et~al.(2007)Joly, Buisson, and Fr{\'e}licot]{joly2007content}
Alexis Joly, Olivier Buisson, and Carl Fr{\'e}licot.
\newblock Content-based copy retrieval using distortion-based probabilistic
  similarity search.
\newblock \emph{ieee Transactions on Multimedia}, 9\penalty0 (2):\penalty0
  293--306, 2007.

\bibitem[Joly et~al.(2003)Joly, Fr{\'e}licot, and Buisson]{joly2003robust}
Alexis Joly, Carl Fr{\'e}licot, and Olivier Buisson.
\newblock Robust content-based video copy identification in a large reference
  database.
\newblock In \emph{International Conference on Image and Video Retrieval},
  pages 414--424. Springer, 2003.

\bibitem[Torralba et~al.(2008)Torralba, Fergus, and Weiss]{torralba2008small}
Antonio Torralba, Rob Fergus, and Yair Weiss.
\newblock Small codes and large image databases for recognition.
\newblock In \emph{2008 IEEE Conference on Computer Vision and Pattern
  Recognition}, pages 1--8. IEEE, 2008.

\bibitem[Jain et~al.(2008)Jain, Kulis, and Grauman]{jain2008fast}
Prateek Jain, Brian Kulis, and Kristen Grauman.
\newblock Fast image search for learned metrics.
\newblock In \emph{2008 IEEE Conference on computer vision and pattern
  recognition}, pages 1--8. IEEE, 2008.

\bibitem[Indyk(2000)]{indyk2000stable}
Piotr Indyk.
\newblock Stable distributions, pseudorandom generators, embeddings and data
  stream computation.
\newblock In \emph{Proceedings 41st Annual Symposium on Foundations of Computer
  Science}, pages 189--197. IEEE, 2000.

\bibitem[Mikolajczyk and
  Schmid(2005{\natexlab{b}})]{mikolajczyk2005performance}
Krystian Mikolajczyk and Cordelia Schmid.
\newblock A performance evaluation of local descriptors.
\newblock \emph{IEEE transactions on pattern analysis and machine
  intelligence}, 27\penalty0 (10):\penalty0 1615--1630, 2005{\natexlab{b}}.

\bibitem[Chum et~al.(2008)Chum, Philbin, Zisserman, et~al.]{chum2008near}
Ondrej Chum, James Philbin, Andrew Zisserman, et~al.
\newblock Near duplicate image detection: min-hash and tf-idf weighting.
\newblock In \emph{BMVC}, volume 810, pages 812--815, 2008.

\bibitem[Auclair et~al.(2009)Auclair, Vincent, and Cohen]{auclair2009hash}
Adrien Auclair, Nicole Vincent, and Laurent~D Cohen.
\newblock Hash functions for near duplicate image retrieval.
\newblock In \emph{2009 Workshop on Applications of Computer Vision (WACV)},
  pages 1--6. IEEE, 2009.

\bibitem[Nister and Stewenius(2006)]{nister2006scalable}
David Nister and Henrik Stewenius.
\newblock Scalable recognition with a vocabulary tree.
\newblock In \emph{Computer vision and pattern recognition, 2006 IEEE computer
  society conference on}, volume~2, pages 2161--2168. Ieee, 2006.

\bibitem[Jiang et~al.(2014)Jiang, Zhang, Li, and Metaxas]{jiang2014computer}
Menglin Jiang, Shaoting Zhang, Hongsheng Li, and Dimitris~N Metaxas.
\newblock Computer-aided diagnosis of mammographic masses using scalable image
  retrieval.
\newblock \emph{IEEE Transactions on Biomedical Engineering}, 62\penalty0
  (2):\penalty0 783--792, 2014.

\bibitem[Khan et~al.(2011)Khan, McCane, and Wyvill]{khan2011sift}
Nabeel~Younus Khan, Brendan McCane, and Geoff Wyvill.
\newblock Sift and surf performance evaluation against various image
  deformations on benchmark dataset.
\newblock In \emph{2011 International Conference on Digital Image Computing:
  Techniques and Applications}, pages 501--506. IEEE, 2011.

\bibitem[Ke and Sukthankar(2004{\natexlab{b}})]{ke2004pca}
Yan Ke and Rahul Sukthankar.
\newblock Pca-sift: A more distinctive representation for local image
  descriptors.
\newblock In \emph{Proceedings of the 2004 IEEE Computer Society Conference on
  Computer Vision and Pattern Recognition, 2004. CVPR 2004.}, volume~2, pages
  II--II. IEEE, 2004{\natexlab{b}}.

\bibitem[Simoncelli and Freeman(1995)]{simoncelli1995steerable}
Eero~P Simoncelli and William~T Freeman.
\newblock The steerable pyramid: A flexible architecture for multi-scale
  derivative computation.
\newblock In \emph{Proceedings., International Conference on Image Processing},
  volume~3, pages 444--447. IEEE, 1995.

\bibitem[He et~al.(2002)He, Zheng, and Ahalt]{he2002object}
Chao He, Yuan~F Zheng, and Stanley~C Ahalt.
\newblock Object tracking using the gabor wavelet transform and the golden
  section algorithm.
\newblock \emph{IEEE transactions on multimedia}, 4\penalty0 (4):\penalty0
  528--538, 2002.

\bibitem[Do and Vetterli(2005)]{do2005contourlet}
Minh~N Do and Martin Vetterli.
\newblock The contourlet transform: an efficient directional multiresolution
  image representation.
\newblock \emph{IEEE Transactions on image processing}, 14\penalty0
  (12):\penalty0 2091--2106, 2005.

\bibitem[Vo et~al.(2006)Vo, Nguyen, and Oraintara]{vo2006texture}
An~Phuoc~Nhu Vo, Truong~T Nguyen, and Soontorn Oraintara.
\newblock Texture image retrieval using complex directional filter bank.
\newblock In \emph{2006 IEEE International Symposium on Circuits and Systems},
  pages 4--pp. IEEE, 2006.

\bibitem[G{\'o}recki et~al.(2012)G{\'o}recki, Sopy{\l}a, and
  Drozda]{gorecki2012ranking}
Przemys{\l}aw G{\'o}recki, Krzysztof Sopy{\l}a, and Pawe{\l} Drozda.
\newblock Ranking by k-means voting algorithm for similar image retrieval.
\newblock In \emph{International Conference on Artificial Intelligence and Soft
  Computing}, pages 509--517. Springer, 2012.

\bibitem[Schmid and Mohr(1997)]{schmid1997local}
Cordelia Schmid and Roger Mohr.
\newblock Local grayvalue invariants for image retrieval.
\newblock \emph{IEEE transactions on pattern analysis and machine
  intelligence}, 19\penalty0 (5):\penalty0 530--535, 1997.

\bibitem[N{\"u}rnberger et~al.(2016)]{nurnberger2016near}
Andreas N{\"u}rnberger et~al.
\newblock Near-duplicate retrieval: a benchmark study of modified sift
  descriptors.
\newblock In \emph{International Conference on Pattern Recognition Applications
  and Methods}, pages 121--138. Springer, 2016.

\bibitem[Xu et~al.(2011{\natexlab{b}})Xu, Ling, Zou, Lu, and Li]{xu2011novel}
Zhihua Xu, Hefei Ling, Fuhao Zou, Zhengding Lu, and Ping Li.
\newblock A novel image copy detection scheme based on the local
  multi-resolution histogram descriptor.
\newblock \emph{Multimedia Tools and Applications}, 52\penalty0 (2-3):\penalty0
  445--463, 2011{\natexlab{b}}.

\bibitem[Alzu’bi et~al.(2016)Alzu’bi, Amira, Ramzan, and
  Jaber]{alzu2016improving}
Ahmad Alzu’bi, Abbes Amira, Naeem Ramzan, and Tareq Jaber.
\newblock Improving content-based image retrieval with compact global and local
  multi-features.
\newblock \emph{International Journal of Multimedia Information Retrieval},
  5\penalty0 (4):\penalty0 237--253, 2016.

\bibitem[Heikkil\"{a} and Pietik\"{a}inen(2006)]{LBP1}
Marko Heikkil\"{a} and Matti Pietik\"{a}inen.
\newblock A texture-based method for modeling the background and detecting
  moving objects.
\newblock \emph{IEEE Trans. on PAMI}, 2006.

\bibitem[Ambai and Yoshida(2011)]{CARD}
Mitsuru Ambai and Yuichi Yoshida.
\newblock {CARD: Compact And Real-time Descriptors}.
\newblock In \emph{ICCV}, 2011.

\bibitem[C. et~al.(2010)C., L., S., and F.]{Michael2010}
Michael C., Vincent L., Christoph S., and Pascal F.
\newblock {BRIEF:} binary robust independent elementary features.
\newblock In \emph{ECCV}, 2010.

\bibitem[Liu et~al.(2019)Liu, Yu, Chen, Li, and Fan]{liu2019top}
Yujie Liu, Deng Yu, Xiaoming Chen, Zongmin Li, and Jianping Fan.
\newblock Top-sift: the selected sift descriptor based on dictionary learning.
\newblock \emph{The Visual Computer}, 35\penalty0 (5):\penalty0 667--677, 2019.

\bibitem[Galshetwar et~al.(2019)Galshetwar, Waghmare, Gonde, and
  Murala]{galshetwar2019local}
GM~Galshetwar, LM~Waghmare, AB~Gonde, and S~Murala.
\newblock Local energy oriented pattern for image indexing and retrieval.
\newblock \emph{Journal of Visual Communication and Image Representation},
  64:\penalty0 102615, 2019.

\bibitem[Zheng and Zheng(2019)]{zheng2019image}
Yi~Zheng and Ping Zheng.
\newblock Image matching based on harris-affine detectors and translation
  parameter estimation by phase correlation.
\newblock In \emph{2019 IEEE 4th International Conference on Signal and Image
  Processing (ICSIP)}, pages 106--111. IEEE, 2019.

\bibitem[Kulkarni et~al.(2019)Kulkarni, Stranieri, and
  Jelinek]{kulkarni2019comparing}
Pradnya Kulkarni, Andrew Stranieri, and Herbert Jelinek.
\newblock Comparing pixel n-grams and bag of visual word features for the
  classification of diabetic retinopathy.
\newblock In \emph{Proceedings of the Australasian Computer Science Week
  Multiconference}, pages 1--7, 2019.

\bibitem[Amato et~al.(2016)Amato, Falchi, and Gennaro]{amato2016reducing}
Giuseppe Amato, Fabrizio Falchi, and Claudio Gennaro.
\newblock On reducing the number of visual words in the bag-of-features
  representation.
\newblock \emph{arXiv preprint arXiv:1604.04142}, 2016.

\bibitem[Philbin et~al.(2007{\natexlab{b}})Philbin, Chum, Isard, Sivic, and
  Zisserman]{philbin2007object}
James Philbin, Ondrej Chum, Michael Isard, Josef Sivic, and Andrew Zisserman.
\newblock Object retrieval with large vocabularies and fast spatial matching.
\newblock In \emph{Computer Vision and Pattern Recognition, 2007. CVPR'07. IEEE
  Conference on}, pages 1--8. IEEE, 2007{\natexlab{b}}.

\bibitem[Sivic and Zisserman(2003)]{sivic2003video}
Josef Sivic and Andrew Zisserman.
\newblock Video google: A text retrieval approach to object matching in videos.
\newblock In \emph{null}, page 1470. IEEE, 2003.

\end{thebibliography}






\end{document}